# Identification of Biased Terms in News Articles by Comparison of Outlet-specific Word Embeddings

Timo Spinde[1,2][0000-0003-3471-4127], Lada Rudnitckaia[1][0000-0003-2444-8056], Felix Hamborg[1][0000-0003-2444-8056], and Bela Gipp[2][0000-0001-6522-3019]

[1] University of Konstanz, Konstanz, Germany
{firstname.lastname}@uni-konstanz.de
[2] University of Wuppertal, Wuppertal, Germany
{last}@uni-wuppertal.de

**Abstract.** Slanted news coverage, also called media bias, can heavily influence how news consumers interpret and react to the news. To automatically identify biased language, we present an exploratory approach that compares the context of related words. We train two word embedding models, one on texts of left-wing, the other on right-wing news outlets. Our hypothesis is that a word's representations in both word embedding spaces are more similar for non-biased words than biased words. The underlying idea is that the context of biased words in different news outlets varies more strongly than the one of non-biased words, since the perception of a word as being biased differs depending on its context. While we do not find statistical significance to accept the hypothesis, the results show the effectiveness of the approach. For example, after a linear mapping of both word embeddings spaces, 31% of the words with the largest distances potentially induce bias. To improve the results, we find that the dataset needs to be significantly larger, and we derive further methodology as future research direction. To our knowledge, this paper presents the first in-depth look at the context of bias words measured by word embeddings.

## 1 Introduction

News coverage is not just the communication of facts; it puts facts into context and transports specific opinions. The way how "the news cover a topic or issue can decisively impact public debates and affect our collective decision making" [12], slanted news can heavily influence the public opinion [11]. However, only a few research projects yet focus on automated methods to identify such bias.

One of the reasons that make the creation of automated methods more difficult is the complexity of the problem: How we perceive bias is not only dependent on the word itself, but also its context, the medium, and the background of every reader. While many current research projects focus on collecting linguistic features to describe media bias, we present an implicit approach to the issue. The main question we want to answer is:

Comparing biased words among word embeddings created from different news outlets, are they more distant (or close) to each other than non-biased words?

To answer this question, we measure any word's context by word embeddings, which reflect the specific usage of a word in a particular medium [15]. We focus on the definition of language bias given by Recasens et al. [10], describing biased words as subjective and linked to a particular point of view. Such words can also change the believability of a statement [10].

Overall, our objectives are to:

1) Analyse and compare the word embeddings of potential bias inducing words trained on different news outlets.

2) Test the assumption that distances between vectors of similar bias words trained on different corpora are larger than between neutral words due to usage in a specific context.

## 2      Related work

While some scholars propose methods to create bias lexica automatically, none of them is in the domain of news articles. Recasens et al. [10] create a static bias lexicon based on Wikipedia bias-driven edits, which they combine with a set of various linguistic features. Ultimately, they aim to classify words as being biased or not. Hube & Fetahu [5] extend this approach by manually selecting bias-inducing words from a highly biased source (Conservapedia) and retrieving semantically close words in a Wikipedia word embedding space.

Since there is no large-scale dataset from which initial knowledge about biased language can be derived, implementation and extending of the approaches of Recasens et al. and Hube & Fetahu may be relevant for news data. However, in the context of media bias identification, creating static bias lexica is inefficient because the interpretation of language and wording strongly depends on its context [3].

It is therefore desirable to either evaluate every word independent of pre-defined lexica or, even more, enable existing biased lexica to be context-aware. In this regard, exploiting the properties of word embeddings is especially interesting [15]. Word embeddings are highly dependent on training corpora they are obtained from and accurately reflect biases and stereotypes in the training corpora [15]. Kozlowski et al. [6] use word embeddings trained on literature from different decades to estimate the evolution of social class markers over the 20th century.

Mikolov et al. [9] compare word embeddings obtained from different languages and show that similar words have minimal cosine similarity. Tan et al. [15] analyze the usage of the same words in Twitter and Wikipedia by comparing their different word representations – one trained on Twitter data and another on Wikipedia.

## 3      Methodology

We seek to devise an automated method that ultimately finds biased words by comparing two (or more) word embeddings spaces, each trained on a differently slanted group

3of text documents. In this exploratory study, we devise a one-time process that consists of four tasks: selection of a word embedding model, selection of biased words, data processing and analysis, and linear mapping of the word embedding spaces.

### 3.1 Word embeddings and parameter selection

To calculate our embeddings, we use Word2Vec [8] with the Continuous Skip-gram (SG) architecture, which achieves better semantic accuracy and slightly better overall accuracy than the Bag-of-Words architecture [9]. We evaluated our word representations via an estimation of word semantic similarity on two datasets – WordSim-353 [2] and MEN [1] and the Google analogy test set [8].

WordSim-353 consists of 353 pairs assessed by semantic similarity with a scale from 0 to 10. MEN consists of 3,000 pairs assessed by semantic relatedness and scaled from 0 to 50. We use these datasets since they focus on topicality and semantics. The Google analogy test set consists of 8,869 semantic and 10,675 syntactic questions. Generally, for our task, the data sets are not ideal, which we discuss in section 5.

We summarize our hyper-parameters in Table 1 and the summary evaluation of our word embeddings in Table 2. We train the word embeddings on the data preprocessed with Genism simple preprocessing and n-grams generated within two passes.

**Table 1.** Hyper-parameters for training the word embeddings

| Hyper-parameter | Value | Hyper-parameter | Value |
| --- | --- | --- | --- |
| dimensionality | 300 | maximum token length | 28 |
| window size | 8 | n-grams threshold (1st pass) | 90 |
| subsampling rate | $10^{-5}$ | n-grams threshold (2nd pass) | 120 |
| # of iterations | 10 | articles titles | included |
| minimum frequency | 25 | training sentence | the whole article |
| function | hierarchical softmax | | |

**Table 2.** Evaluation of the word embeddings

| Corpora | # articles | # tokens | Vocabulary size | Semantic similarity | | Analogy |
| --- | --- | --- | --- | --- | --- | --- |
| | | | | WordSim-353 | MEN | Google |
| HuffPost | 101K | 68M | 53K | 0.65 | 0.71 | 0.50 |
| Breitbart | 81K | 39M | 37K | 0.57 | 0.59 | 0.38 |

### 3.2 Manual selection of bias inducing words

We follow the approach proposed by Hube & Fetahu [5] and manually select a small set of "seed" words that are very likely to be related to controversial opinions. They also have a high density of bias-inducing words surrounding them in the embedding space. The 87 seed words are selected based on the description of controversial left and right topics on Allsides.com (Table 3, see also https://www.allsides.com/media-bias/left, …/right). From the list of the closest twenty words to each seed word, we manually extracted words that convey a strong opinion [5]. As the identification of bias is not trivial for humans [10], we validated the extended seed words by four student volunteers, age 23 - 27, who labeled each word as being biased or not. We discarded any words where less than three students agreed on.

**Table 3.** The seed words that are likely to be related to controversial opinions and to have a high density of bias-inducing words surrounding them in the word embedding space

| Divisive issue | Seed words |
| --- | --- |
| The role of the government | regulation(s), involvement, control, unregulated, government, centralization, law |
| Economics | tax(es), taxation, funding, spending, corporation(s), business(es), economy |
| Equality | equality, inequality, rights, equal_rights, wealth, living_wage, welfare, welfare_state |
| Social services | services, government_services, social_security, benefit(s), help, student(s), loan(s), student_loan(s), education, healthcare, individual, personal_responsibility, collective |
| Security | security, military, military_force, defense, intervention, protect, protection, border, border_security, migration, migrant(s), immigration, immigrant(s), terror, terrorist(s) |
| Traditions, religion, and culture | tradition, norms, cultural_norms, progress, change(s), race, racism, gender, sexual, orientation, sexual_orientation, identity, religion, Islam, tolerance, multiculturalism, values, family_values, bible, constitution |
| Miscellaneous | freedom, speech, freedom_of_speech, free_speech, hate_speech, gun(s), gun_owner(s), abortion, environment, media |

### 3.3 Data

We choose two news outlets that are known to take different views and potentially use different words to describe the same phenomena. We based the choice of news outlets for analysis on the media bias ratings provided by Allsides.com. The news aggregator aims to estimate the overall slant of an article and a news outlet by combining users' feedback and expert knowledge [3, 13]. We choose The HuffPost as a left-wing news outlet and Breitbart News as right-wing. We scraped articles from both news outlets, published in the last decade, from 2010 to 2020, from Common Crawl [4]. For preprocessing, we use Genism simple preprocessing and generate n-grams.

### 3.4 Linear mapping between vector spaces

Since the goal is to compare word vectors between two different word embedding spaces, it is necessary to make sure that these two word embedding spaces have similar dimensionality. We use the approach proposed by Mikolov et al. [15] and Tan et al. [20].

The results of training two different mapping matrices – trained on 3,000 most frequent words and on the whole common vocabulary – are presented in Table 4. The only metric to evaluate mapping quality is the number of distant words. Ideally, similar words should be close to each other after linear mapping. A large number of very distant words can be an indicator of a poorly trained matrix.

We assessed the distance between words in an embedding space with cosine similarity. In case the distance between similar words after mapping from one source to another depends on the frequency of the word in either a left- or right-wing source, according to Tan et al. [15], adjusted distances should be compared. The larger an adjusted distance, the less similar the word is between the two sources since positive adjusted distance values belong to words that are less similar than at least half of the words in their frequency bucket.

**Table 4.** Comparison of two linear mappings

| Mapping matrix trained on | # tokens in common vocab. | Median cos. sim. | # distant words | | # close words | Correlation of cos. sim. with freq. in | |
|---|---|---|---|---|---|---|---|
| | | | cos. sim. $\leq 0.4$ | adj. cos. sim. $\geq 0.1$ | cos. sim. $\geq 0.6$ | HuffPost | Breitbart |
| 3K | 30K | 0.48 | 8K (25%) | 5K (17%) | 6K (20%) | 0.14 | 0.13 |
| whole vocab. | | 0.56 | 2K (6%) | 4K (14%) | 10K (35%) | 0.12 | 0.12 |

For both variants, we still obtained many distant words after linear mapping, i.e., median cosine similarity is 0.48 and 0.56 for the first and the second variant, respectively. Ideally, similar words should be close to each other after linear mapping, except those used in different contexts. Possible reasons for having many distant words are:
- low quality of trained word embeddings and, thus non-stable word vectors,
- low quality of mapping matrix, possibly nonlinear transformation is needed,
- a high number of "bad" n-grams,
- a high number of noisy words.

We tried to address the first two causes by training different word embeddings models and different mappings. Since Tan et al. [15] did not discuss the threshold for the definition of distant words, we choose the thresholds to define distant words as lower than 0.4 and higher than 0.1 for pure and adjusted cosine similarities, respectively. Comparing n-grams and unigrams based on their cosine similarity statistics, we conclude that there is no apparent reason to think that the generated n-grams are more distant than the unigrams: median cosine similarity for n-grams is 0.62, whereas for unigrams it is 0.55. We manually inspected the distant words to estimate the possible influence of flaws in preprocessing and connection to bias words (Section 4.1).

The matrix trained on the whole vocabulary maps similar words better since there are fewer very distant words, and the median cosine similarity for the words is higher. Therefore, we used this mapping for further analysis.

At the two-dimensional graph obtained by reducing the dimensions of the word vectors from 300 to 2 with PCA, it can be seen that the mapping works quite well for the most frequent words, here, the pronouns: the word vectors mapped from HuffPost to Breitbart are indeed closer to the vectors from Breitbart than the initial vectors from the HuffPost (Figure 1 a). However, we also get a high number of distant words (Figure 1 b). We can also see that the higher the frequency, the higher the chance that the words are better mapped from one source to another (Figure 1 c). Simultaneously, for less frequent words, the results of mapping vary: some words are mapped very well and some very poorly (Figure 1 d). We can see the same patterns can for adjusted distances in Figure 1, e-h.

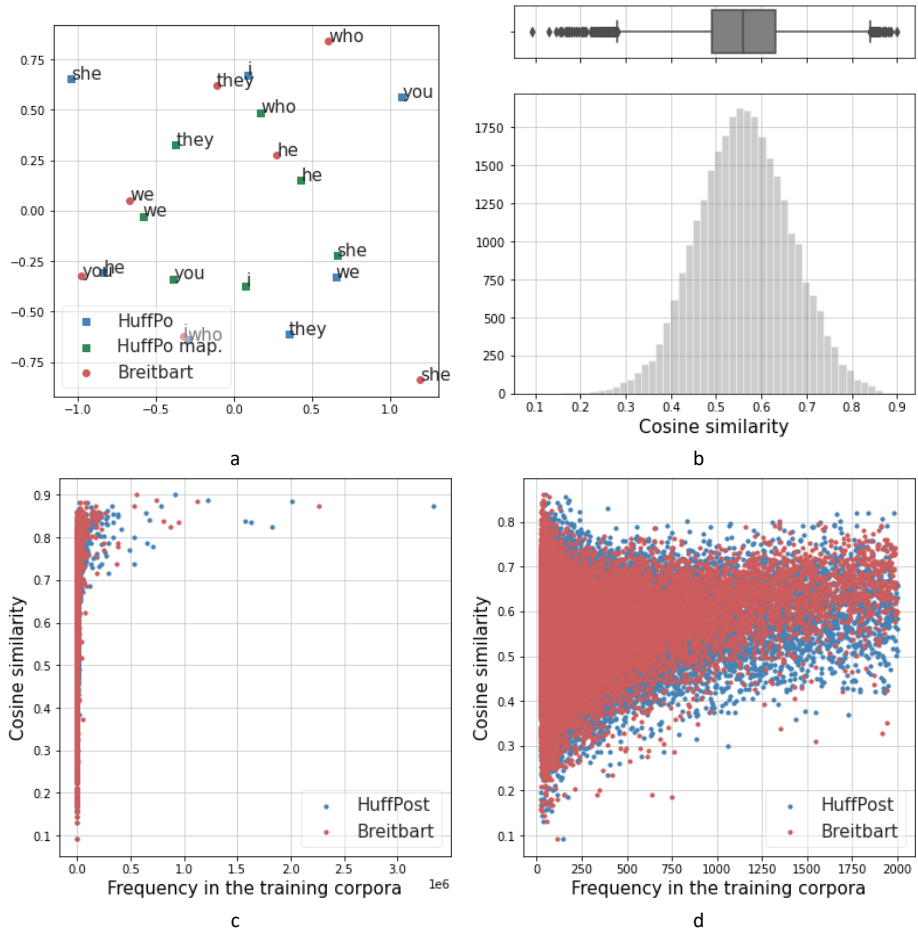

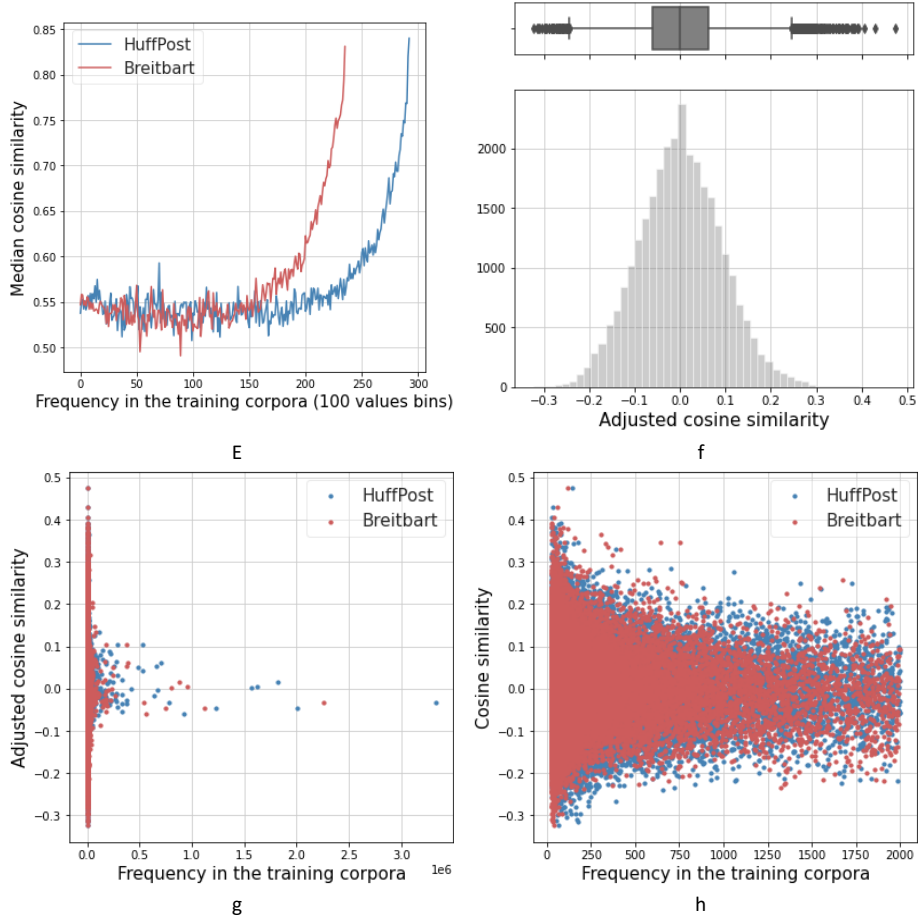

**Fig. 1.** Linear mapping from HuffPost to Breitbart trained on the whole vocabulary: a) high-frequency word vectors before and after mapping, b) distribution of cosine similarities after mapping, c) dependency of frequency and cosine similarities, d) dependency of frequency and cosine similarities for the words less frequent than 2K, e) median cosine similarities per frequency bucket, f) distribution of adjusted cosine similarities after mapping, g) dependency of frequency and adjusted cosine similarities, h) dependency of frequency and adjusted cosine similarities for the words less frequent than 2K.

## 4 Results

### 4.1 Distant words

We manually examine the top 1,000 most distant words, both with low cosine similarity and high adjusted cosine similarity. The lists highly overlap, i.e., the top 1,000 words with high adjusted cosine similarity introduce only 126 new words.

Names and surnames make up a large part of the 1,000 most distant words with 387 occurrences. Fifty-nine words are short words that are either abbreviations or noise. Among the rest, the number of words that can potentially induce bias is 172 (31%), if not to consider mentioned names and short words.

### 4.2 Relation of distant words to bias inducing words

For the manually selected bias words, we find no pattern regarding the distance of their vectors between HuffPost and Breitbart. The median cosine similarity is 0.60, which is slightly higher than the median cosine similarity of the words in the whole common vocabulary. Median adjusted cosine similarity is -0.05, which also shows that the words in this group are, in general, even slightly closer to each other than the words in the same frequency buckets.

Since the number of manually selected words is quite small, we also check the relation of distant words to the words from the bias lexicon automatically created by Hube & Fetahu [9]. Out of 9,742 words in the lexicon, we encountered 3,334 in the common vocabulary of the HuffPost and Breitbart. This check contradicts with the main idea – that bias words differ from context to context and from outlet to outlet – but we conduct it for additional insight and confirmation that bias inducing words are not directly related to distant ones.

Similarly to the manually selected bias words, for the words from the bias lexicon, we found no pattern regarding the distance of their vectors between different outlets. The median cosine similarity is 0.52, slightly lower than the median for all the words in the common vocabulary. The Median adjusted distance is 0.03, which means that the words in this group are, in general, just slightly more distant than other words with the same frequency. Therefore, this finding does not allow to claim that bias words are in general more distant than other words but rather corroborates that bias inducing words are not directly connected with distant words.

Overall, there are no salient differences when comparing the context of biased words between HuffPost and Breitbart. The most noticeable differences are between the context of the words "regulations," "welfare," "security," "border," "immigration," "immigrants," "hate_speech," and "abortion". We also notice the differences in the context of the words that have more than one meaning, e.g., the word "nut" is surrounded by the words describing food in the word embeddings trained on the HuffPost corpus. In contrast, in the word embeddings trained on the Breitbart corpus, it is surrounded by such words as "horrid", "hater", etc. Such findings are rare, and their statistical significance should be proved on the exhaustive biased words lexicon and the word embeddings trained on larger datasets.

## 5 Conclusion and future work

We present experimental results of an approach for the automated detection of media bias using the implicit context of bias words, derived through fine-tuned word embeddings. Our key findings are:

1) Among the words with large distances after linear mapping, some can potentially induce bias. Their percentage is around 25% (if not to consider names, surnames, and short words that can be either abbreviations or noise, otherwise the ratio is about 15%).
2) In the small set of manually selected bias inducing words, median cosine similarity after the linear mapping is 0.6 which is even slightly higher than for the whole vocabulary. A direct relation to large distances also did not show on the words from the bias lexicon provided by Hube et al. [5].
3) There are no salient differences in the context of seed words apart from several words.

Obtained results are either point to the absence of a relation of bias and distant words or can be explained by the following flaws of the current project, which serve as future research directions. First, the data for training our word embeddings are relatively scarce. Intrinsic evaluation of the word embeddings trained on the Breitbart corpora shows low results. Our current evaluation methods do not reflect the actual suitability of the word embeddings for our specific task. Second, we did not test other word embedding models than Word2Vec, which might show a better overall performance, e.g., GloVe, BERT, Elmo, and Context2Vec [7]. We did also not integrate other features, such as lexical cues or sentiment. Third, bias inducing words are selected manually by a tiny group of non-native English speakers. Fourth, we based the comparison of context on the top 20 most similar words. But among these top twenty for one source, the similarity can be on average high and for another on average low.